\documentclass[11pt]{article}

\usepackage[preprint]{acl}

\usepackage{times}
\usepackage{latexsym}

\usepackage[T1]{fontenc}

\usepackage[utf8]{inputenc}

\usepackage{microtype}

\usepackage{inconsolata}

\usepackage{graphicx}

\usepackage{hyperref}
\usepackage{url}
\usepackage{booktabs}
\usepackage{graphicx}
\usepackage{wrapfig}
\usepackage{subcaption}
\usepackage{booktabs}
\usepackage{multirow}
\usepackage{lineno}
\usepackage{enumitem}
\usepackage{longtable}
\usepackage[raster, most]{tcolorbox}

\definecolor{darkblue}{rgb}{0, 0, 0.5}
\hypersetup{colorlinks=true, citecolor=darkblue, linkcolor=darkblue, urlcolor=darkblue}

\newenvironment{promptbox}[1][]{
  \begin{tcolorbox}[
    enhanced,
    breakable,
    colback=gray!10,
    arc=0pt,
    boxrule=0pt,
    title=\scriptsize #1,
    fontupper=\tiny,
  ]
}{
  \end{tcolorbox}
} 

\usepackage{xcolor}
\usepackage{geometry}
\usepackage{array}

\usepackage{xspace}

\title{Strong Memory, Weak Control: 

An Empirical Study of Executive Functioning in LLMs}

\author{
\parbox{\linewidth}{
Karin de Langis\textsuperscript{1}, Jong Inn Park\textsuperscript{1}, Bin Hu\textsuperscript{1}, Khanh Chi Le\textsuperscript{1}, Andreas Schramm\textsuperscript{2}, Michael C. Mensink\textsuperscript{3}, Andrew Elfenbein\textsuperscript{4}, \& Dongyeop Kang\textsuperscript{1} }\\
\textsuperscript{1}Department of Computer Science and Engineering, University of Minnesota \\
\textsuperscript{2}Department of Linguistics, Hamline University \\
\textsuperscript{3}Department of Psychology, University of Wisconsin-Stout \\
\textsuperscript{4}Department of English, University of Minnesota \\
\texttt{\{dento019,park2838,hu000562,le000422,elfen001,dongyeop\}@umn.edu,}\\\texttt{a.schramm@hamline.edu, mensinkm@uwstout.edu} \\
}\author{
\parbox{\linewidth}{
Karin de Langis\textsuperscript{1}, Jong Inn Park\textsuperscript{1}, Bin Hu\textsuperscript{1}, Khanh Chi Le\textsuperscript{1}, Andreas Schramm\textsuperscript{2}, Michael C. Mensink\textsuperscript{3}, Andrew Elfenbein\textsuperscript{4}, \& Dongyeop Kang\textsuperscript{1} }\\
\textsuperscript{1}Department of Computer Science and Engineering, University of Minnesota \\
\textsuperscript{2}Department of Linguistics, Hamline University \\
\textsuperscript{3}Department of Psychology, University of Wisconsin-Stout \\
\textsuperscript{4}Department of English, University of Minnesota \\
\texttt{\{dento019@umn.edu}\\
}

\begin{document}
\maketitle
\begin{abstract}
Working memory, or the ability to hold and manipulate information in the mind, is a critical component of human intelligence and executive functioning.
It is correlated with performance on various cognitive tasks, including measures of fluid intelligence, which encompasses reasoning and problem solving. 
We use a comprehensive set of classic working memory tasks to estimate the working memory capacity of large language models (LLMs).
We find that in most cases, LLMs exceed normative human scores. 
However, we do not find that the increased capacity of working memory is associated with higher performance on other executive functioning tasks or problem solving benchmarks. 
These results suggest that LLMs may have deficits in attentional control and cognitive flexibility, which result in difficulties with inhibiting automatic responses and adapting to shifting information.
Our findings suggest that reasoning models, although they often do not currently fully compensate for these deficits, may have the potential to do so in the future.
\end{abstract}

\section{Introduction}

Working memory is the cognitive function that handles storage and manipulation of task-relevant data. Many tasks, from calculating a tip to playing a complex game of chess, require keeping relevant information in mind and mentally manipulating that information, and thus rely on working memory. Working memory is correlated with fluid intelligence and performance on numerous cognitive tests \cite{conway2013working}, and it has been suggested to be ``perhaps the most significant achievement of human mental evolution'' \cite{goldman1992working}.

Naturally, working memory is also of interest in the study of artificial intelligence \cite{yang2018dataset}
and more recently, in large language models (LLMs) \cite{gong2024working, zhang2024working}. 
Working memory in LLMs has been assessed via the $n$-back test, with results suggesting LLMs have human-like working memory. However, $n$-back is only one of several standard measures of working memory capacity, and it is often not the most representative metric \cite{frost2021n}. 

\begin{figure}[t]
    \centering
    \includegraphics[width=\linewidth]{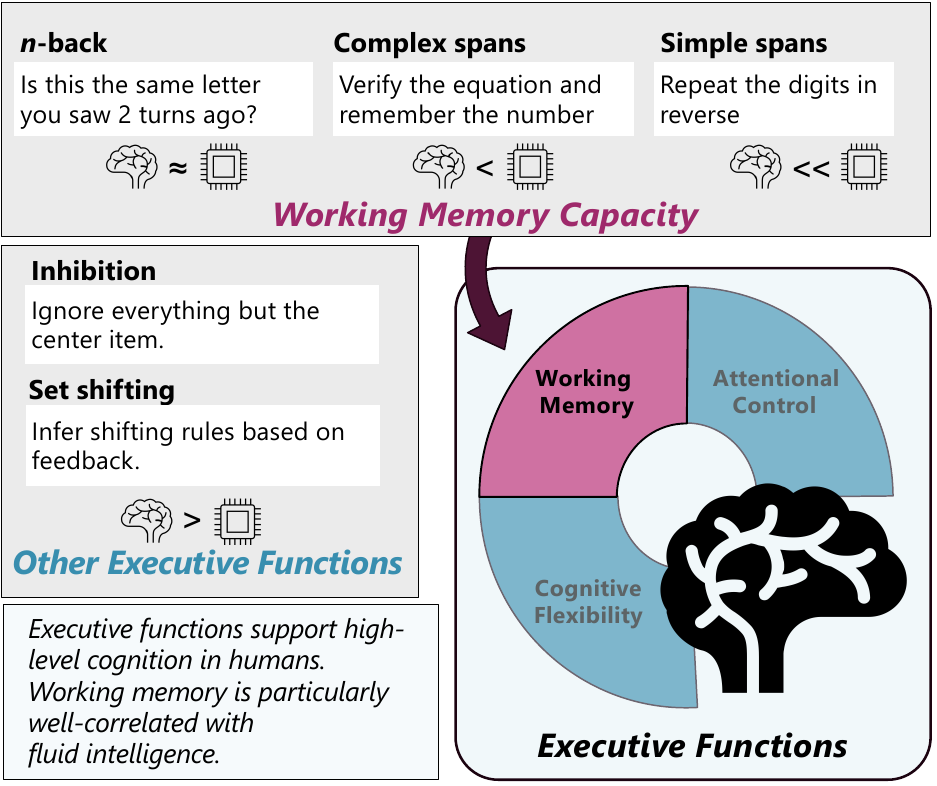}
    \caption{Working memory is a keystone component of high-level human cognition. We administer a standard set of working memory tasks to LLMs and find that they tend outperform humans, especially in the simplest tasks. However, we find that in tasks of executive function, the LLMs were generally below human baselines.
    }
    \label{fig:figure1}
\end{figure}

We propose a more robust estimate of LLM working memory capacity by evaluating LLMs on a full suite of working memory tasks adapted from human studies, taking measures of simple and complex spans in addition to the $n$-back task.
We also extend the study of working memory into executive functions more broadly. Executive functions are a group of high-level cognitive processes that enable sophisticated thinking \cite{diamond2013executive}. There are three primary executive functions: (1) working memory, (2) inhibition / attentional control, and (3) cognitive flexibility. These functions, although correlated with one another, are also clearly separable based on both behavioral and neurological data \cite{miyake2000unity}. We choose one test of attentional control and one test of cognitive flexibility to assess LLM executive functions beyond working memory.\footnote{Specifically, we administer the Eriksen flanker task and the Wisconsin Card Sorting Task (WCST).}
Our experiments consider multiple model families, sizes, and prompts, and we compare both reasoning and instruction tuned models. 

Our findings show that LLMs on most tasks have a substantially higher working memory capacity than humans do, and that this capacity is correlated with model size. Still, LLMs do have relative performance reduction on complex working memory measures that have additional processing or manipulation demands. Since the transformer architecture offers much stronger information encoding than the biological brain, it is not clear whether advanced working memory capacity in LLMs will correlate to better performance in other executive functions. 
Our tests of other executive functions show no human-like correlation with working memory capacity, and further, we find that LLM performance on these tasks is \textit{below} published normative data from human research. This suggests that \textit{the bottleneck for LLM intelligence is not working memory, but other critical executive functions required to perform higher-order tasks}.

Finally, we examine the performance of reasoning model on these cognitive tasks, since the thought strings they produce could theoretically take the role of the central executive by carrying out selection and goal-oriented manipulation of task relevant information.  
We find mixed results in performance benefits: they improve performance on a single-turn attentional control task, but not on a multi-turn cognitive flexibility task. We also consistently observe the overthinking phenomenon \cite{chennot} in which simple tasks result in excessively long thought strings, making performance very inefficient. 
We hypothesize that \textit{with specific cognitive task training, reasoning models may learn to consistently serve as an artificial executive function mechanism}, which may in turn open the door to more intelligent LLMs.

\section{Related Work}\label{sec:rel}
Working memory is among the most studied cognitive ability in humans, perhaps because it has been correlated with a large number of cognitive skills. In particular, working memory plays an important role in executive function and problem solving, and thus has a high impact on goal-oriented behavior and critical thinking. For an excellent overview on the relation between working memory and intelligence, see \citet{unsworth2014working}.

Working memory is a temporary information store with limited capacity \cite{baddeley1994developments, cowan2001magical}. There are two critical components of working memory. The first is the maintenance or storage of relevant information -- i.e., keeping the information ``in mind'' --  and the second is the transformation, processing, or updating of that information as needed. 
Neurological studies indicate that the \textit{maintenance} of information in working memory is enabled by sustained activations of neural circuits \cite{goldman1995cellular}. The neural substrates behind the manipulation of this information are not yet well understood. 

The study of working memory in LLMs is still developing. \citet{gong2024working} found similar performance between LLMs and humans on the n-back task, a common assessment of working memory. Later work further investigates possible prompting strategies to enhance model performance on the task \cite{zhang2024working}. Finally, drawing inspiration on cognitive difficulties associated with overloading working memory in humans, \citet{upadhayay2025working} demonstrate that overloading LLMs with unrelated tasks is an effective jailbreaking technique, suggesting that LLMs may also have a limited working memory capacity. 

Other types of memory\footnote{Note that working memory is functionally dissociable from other forms of memory, as evidenced by double dissociation in patients with focal brain lesions.} have a longer history of study in LLMs. Long term memory, or the background knowledge learned during pre-training, is well-established \cite{petroni2019language}; the formation and forgetting of long-term memory over the course of pre-training has been investigated \cite{chang2024large} and precise weights storing facts have been localized and causally demonstrated to facilitate factual recall \cite{de2021editing}. Exact sequence memorization during training has also been extensively studied, with theoretical bounds for sequence lengths conditioned on the number of parameters in a transformers-based architecture being established \cite{kim2023provable, kajitsukaoptimal, yuntransformers, zaheer2020big}. The associative memory capability of transformers has also been theoretically established \cite{jiang2024llms}, and in-context memorization has been compared to declarative memory in humans \cite{li2024linking} and linked to the transformers architecure \cite{akyureklearning}. \citet{jelassi2024repeat} show that the transformers architecture excels at repeating in-context sequences, which strongly resembles a short-term memory store in humans (also \cite{wenrnns}). Other features of human memory, such as recency effects and priming effects, have been documented in LLMs as well \cite{janik2023aspects, jumelet2024language, Shaki_2023}. Our work expands the study of working memory in LLMs, with particular focus on whether LLM working memory modulates downstream skills as it does in humans.

\section{Experimental Setup}

\begin{figure}[t]
    \centering
    \includegraphics[width=\linewidth, trim=0.2cm 0.2cm 0.2cm 0.2cm,clip]{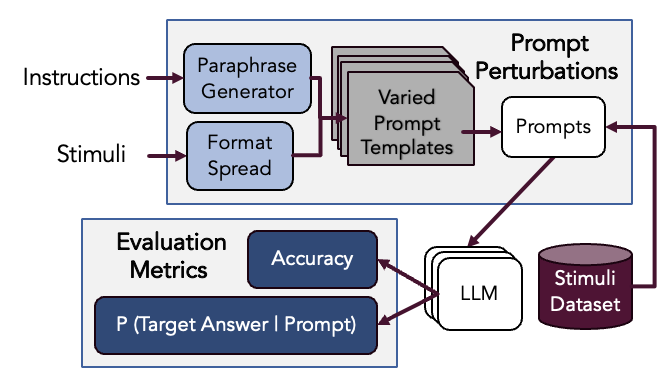}
    \caption{Our experimental pipeline.}
    \label{fig:pipeline}
\end{figure}

Since cognitive processes cannot be directly observed and individual results may reflect task-specific artifacts, cognitive science aims to collect observations across diverse tasks and populations.
Following this as the primary design principle, we introduce variations in our experimental setups (Figure \ref{fig:pipeline}): models, prompt templates, and task stimuli to see if we obtain comparable findings under different conditions.

\paragraph{Prompt Variations. }\label{sec:paraphrasing}
When evaluating LLM responses to cognitive tasks, it is more desirable to marginalize over all prompts for a given task, rather than relying on one prompt to evaluate performance, due to high prompt sensitivity in LLM responses \citep{sclar2024quantifying, wahle-etal-2024-paraphrase}. We create thirty prompt permutations based on \textsc{FormatSpread} \cite{sclar2024quantifying} and paraphrasing of instructions (Figure~\ref{fig:pipeline}). For details, see \S\ref{sec:prompts}. We establish that these prompts induce spread in model accuracy as expected (see \S\ref{sec:prompt_spread}).

\paragraph{Model Variations.}
We use six open-source LLMs (instruction-tuned variants) from three different model families, each with two size variations: Gemma2 with 9B and 27B parameters \citep{team2024gemma}, Llama3.1 with 8B and 70B parameters \citep{dubey2024llama}, and Qwen2 with 7B and 72B parameters \citep{yang2024qwen2}. The smaller models additionally vary in their number of attention heads, which is theoretically relevant to in-context memorization: Qwen2-7B has 28 attention heads, Gemma2-9B has 16, and Llama-3.1-8B has 32. We also test reasoning variants of two models: a DeepSeek-R1 \citep{deepseekai2025deepseekr1} distilled Llama3.1-8B  and Qwen3-8B. We use the Huggingface library \citep{wolf-etal-2020-transformers} for model inference, applying 4-bit quantization for larger models (27B or more parameters) to meet computational constraints. For experiments with long contexts, we also use Flash Attention \citep{dao2022flashattention}. Otherwise, default model configurations and generation hyperparameters are used in all cases. All experiments are completed on a Linux server equipped with two Nvidia A100 GPUs.



\begin{figure*}[ht]
    \centering
    \begin{subfigure}[b]{0.49\linewidth}
        \centering
        \includegraphics[width=\linewidth,trim=0 0.7cm 0 0, clip]{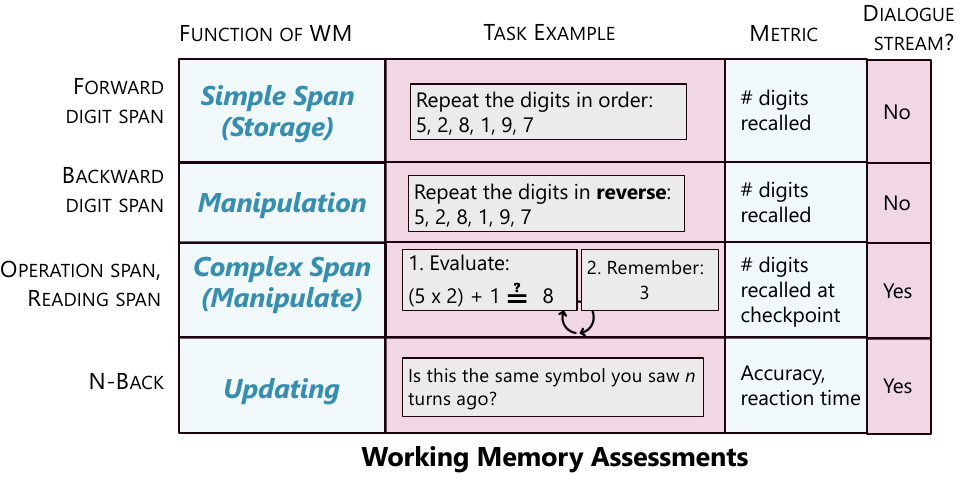}
        \caption{Working memory assessments}
        \label{fig:wmoverview}
    \end{subfigure}
    \begin{subfigure}[b]{0.49\linewidth}
        \centering
        \includegraphics[width=\linewidth,trim=0 0.8cm 0 0, clip]{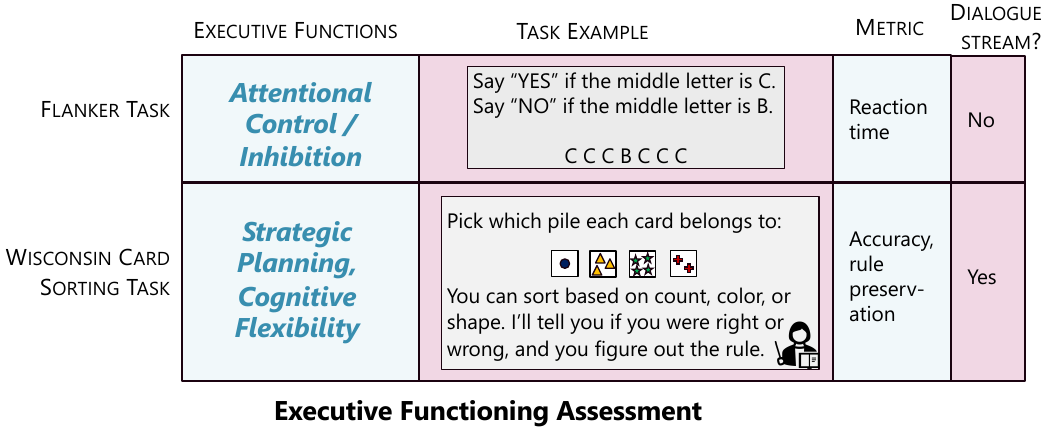}
        \caption{Executive function assessments}
        \label{fig:sub2}
    \end{subfigure}

    \caption{Brief overview of the cognitive assessments adapted for LLMs in the present study: (a)  working memory tasks, spanning from simple input/output pairs to full dialogues, (b) executive function tasks, ranging relatively simple flanker task to complex Wisconsin Card Sorting.
    }
    \label{fig:task_overview}
\end{figure*}

\paragraph{Task Variation}

\begin{figure}[t]
    \centering
    \includegraphics[width=\linewidth]{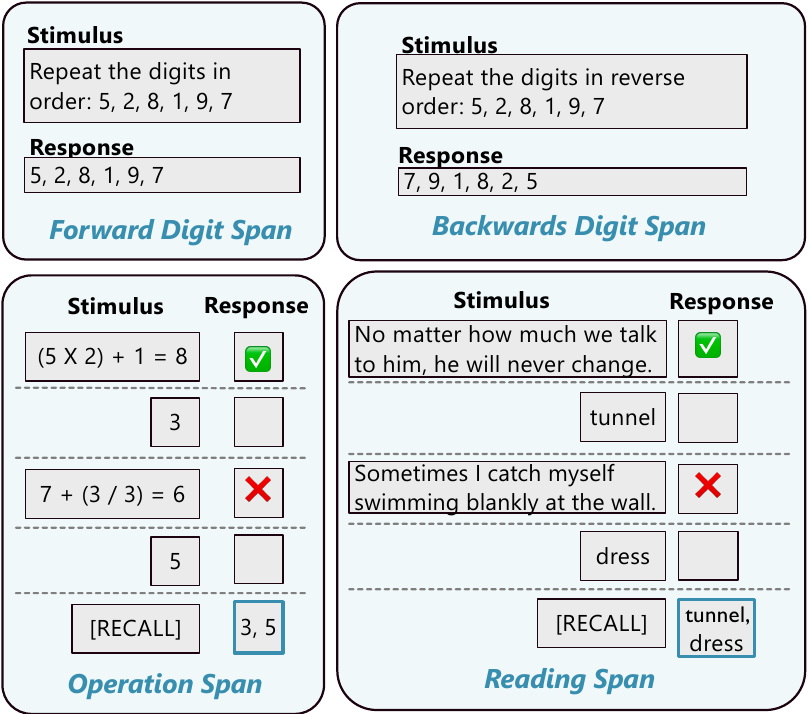}
    \caption{We asses both \textbf{simple span tasks} (top) and \textbf{complex span tasks} (bottom). The simple forward digit span task assesses the storage component of short term memory, whereas backward digit span has some overlap with working memory due to the manipulation component. On the other hand, complex span tasks also heavily tax information processing, making them a frequent choice to estimate of working memory spans.
    }
    \label{fig:spans}
\end{figure}

The following result sections answer three research questions, each addressed by a different set of tasks:  
\begin{itemize}[left=3pt, itemsep=1em, topsep=5pt, parsep=1pt]
    \item \textbf{RQ1 (\S\ref{sec:wmc}, \S\ref{sec:wmc_complex}):} Do LLMs show human-like working memory capacity?  
    We measure access and manipulation limits in LLMs from five working memory tasks (Figure~\ref{fig:spans}): two simple spans, two complex spans, and the n-back task. 
    \item \textbf{RQ2 (\S\ref{sec:cognitive}):} Does working memory capacity in LLMs transfer to other executive functions?  
    To assess executive functions, we measure tasks linked to attentional control and strategic flexibility: one simple task (flanker) and one complex task (WCST) (Figure \ref{fig:wcst_timeline}). These tasks allow us to estimate both LLM working memory capacity and whether it supports broader cognitive abilities, i.e., executive functions.
    \item \textbf{RQ3 (\S\ref{sec:reasoning}):} Can reasoning models provide cognitive benefits that can supplement LLM cognition?  
    We assess reasoning effects on working memory capacity (\S\ref{sec:reasoning}) and the cognitive tasks from RQ2. 
\end{itemize}

\section{Estimating working memory capacity}\label{sec:estimating_wmc}
In this section, we detail experiments assessing working memory capacity in LLMs (\S \ref{sec:wmc}, \S\ref{sec:wmc_complex}), followed by tests of other executive functions \S\ref{sec:cognitive}.
Finally, we discuss results obtained from reasoning models (\S\ref{sec:reasoning}).

\subsection{LLMs generally have larger working memory spans than humans}\label{sec:wmc}
Standard measures of working memory capacity are the backward digit span, operation span, and reading span (see Figure~\ref{fig:spans}) and the $n$-back task.
Humans struggle with these tasks because stimuli vanish instantly, subject to rapid decay, and easily disrupted by interference. Transformers, by contrast, do not suffer from sensory decay and can reliability reproduce long sequences with their context windows by leveraging the self-attention mechanism. \cite{jelassi2024repeat}. 

Prior work shows that LLMs have human-like $n$-back performance \cite{gong2024working}. To assess LLM working memory capacity more robustly, we apply a battery of classic tasks used in psychology. (For details on specific prompting and evaluation set up, see \S\ref{sec:prompts}.) We reproduce \citet{gong2024working} results, but for all other working memory tasks, \textit{LLMs meet or surpass human baselines} (Table~\ref{tab:wmc}).
Moreover, we find parameter count correlates strongly with mean span performance ($r = 0.82, p < 0.05$), while the number of attention heads shows a weaker trend ($r = 0.76; p = 0.08$).
This suggests that while attention heads may contribute to straightforward repetition (e.g., forward digit span), they are \textit{less predictive of tasks requiring manipulation}, such as backward or flexible repetition. 

\begin{table*}[h!]
\centering
\small
\begin{tabular}{lccccccc}
\toprule
\textbf{Model} & \textbf{1-back} & \textbf{2-back} & \textbf{3-back} & \textbf{O-SPAN} & \textbf{R-SPAN} & \textbf{BDS ($d = 15$)} & \textbf{FDS ($d = 50$)} \\ 
\midrule
Gemma-2-9B     & 0.99 & 0.75 & 0.72 & 0.93 & 0.97 & 0.21 & 0.99 \\
Gemma-2-27B    & 0.91 & 0.72 & 0.69 & 0.92 & 0.98 & 0.59 & 1.00 \\
Llama-3.1-8B   & 0.76 & 0.68 & 0.67 & 0.99 & 0.92 & 0.18 & 1.00 \\
Llama-3.1-70B  & 0.93 & 0.82 & 0.82 & 0.99 & 0.94 & 0.83 & 1.00 \\
Qwen2-7B       & 0.99 & 0.89 & 0.85 & 0.96 & 0.66 & 0.00 & 1.00 \\
Qwen2-72B      & 0.78 & 0.74 & 0.70 & 0.93 & 0.97 & 0.51 & 1.00\\
\midrule
Human (approx) & 0.98 & 0.91 & 0.75 & 0.53 & 0.48 & 0.00 & 0.00\\
\bottomrule
\end{tabular}
\caption{Average model accuracy on each test. Rough estimates of typical human scores are provided for reference (see \S\ref{sec:human-baseline}). Note that for the digit span tasks, we include accuracies for very long strings ($d = 15$ and $d = 50$), while the typical human span is 5 (BDS) to 7 (FDS).}\label{tab:wmc}
\end{table*}

\subsection{LLMs excel at simple spans but concurrent processing in complex spans induces more errors}\label{sec:wmc_complex}
Memory span tasks are often grouped into two categories: simple spans, i.e. forward and backward digit spans, and complex spans, i.e. operation and reading spans \cite{kane2003working}. In Table~\ref{tab:wmc}, we see that LLMs vastly exceed human performance on simple spans, but are less dramatically ahead in complex spans, indicating that additional information processing demands can hinder LLM information retrieval in a human-like fashion. In a similar vein, consider LLM performance on the forward and backward digit spans. LLMs are perfect up to 50 forward digits, but already begin to falter at only 15 backward digits. The reversal operation seems to interfere with recall ability. These findings suggest that while LLMs' architecture supports large storage capacity, \textit{their ability to concurrently integrate information processing and memory is more limited}.
In Section~\ref{sec:reasoning}, we examine whether reasoning models can better handle these complex working memory demands.


\subsection{LLMs are below human baselines on tests of other executive functions} \label{sec:cognitive}
Human working memory capacity is correlated to a wide range of cognitive abilities, including other executive functions \cite{unsworth2014working}.
Executive functions are a suite of high-level cognitive skills that support sophisticate goal-driven behavior. 
Having found that LLMs exceed human baselines of working memory capacity (\S~\ref{sec:wmc}), we next ask whether this advantage extends to other executive functions, which rely on working memory but require additional control and flexibility. 

\begin{figure*}
    \centering
    \includegraphics[width=0.98\linewidth]{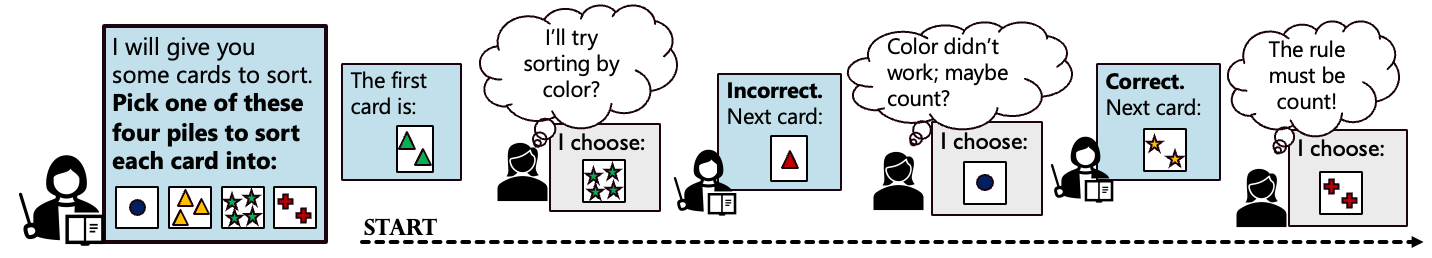}
    \caption{The Wisconsin Card Sorting Task (WCST). Participants are serially presented with cards to sort, and they must infer an underlying ``sorting rule'' based on feedback. \textit{The rule will periodically change without warning}, and people must detect the change and adjust to the new rule. Participants make sorting decisions as quickly as possible. The WCST tests various executive functions, primarily cognitive flexibility as participants adapt from one set of rules to another.}
    \label{fig:wcst_timeline}
\end{figure*}

\subsubsection*{LLMs show low aptitude on WCST}
We first test LLMs on the Wisconsin Card Sorting Task (WCST) \citep{grant1948behavioral}, a widely studied measure of cognitive flexibility, abstract reasoning, and set-shifting \citep{miles2021considerations}. In this task, participants must infer and apply a sorting rule from feedback, and then adapt when the rule changes without warning (Figure~\ref{fig:wcst_timeline} shows a schematic).
Performance on this task is moderately correlated with working memory capacity on complex spans \cite{lehto1996executive, dann2023measuring}.
We translate the WCST to a textual format for LLMs.\footnote{We also create a visual version and test on VLMs but find that performance is better on the textual version. For details and results on the VLM study see \S\ref{sec:wcst-vlmodels}.} Each model receives 102 serially presented stimuli, with three examples illustrating each sorting rule. 

In human studies, instructions typically that the sorting rule must be inferred from feedback, and that the rule may change. After pilot studies with human-like instructions showed LLMs were unable to establish an effective strategy under these conditions (\S\ref{sec:WCST-appendix}), we adjusted prompts to also specify an exact strategy to apply in response to learning a guess was correct or incorrect based on feedback. Full prompting and stimuli presentation details are in \S\ref{sec:prompts}. 

\begin{figure}[t]
    \centering
    \begin{subfigure}[b]{0.48\linewidth}
        \includegraphics[width=\linewidth]{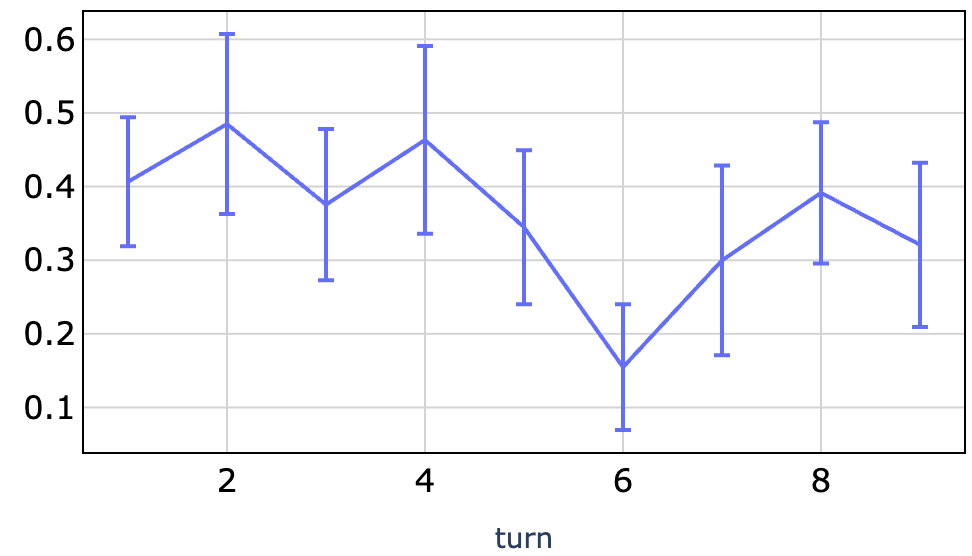}
        \caption{Gemma-2-9B: Frequency of correct answers for turns 1 - 9 after seeing a new rule.}
    \end{subfigure}
    \hfill
    \begin{subfigure}[b]{0.48\linewidth}
        \includegraphics[width=\linewidth,trim={0 0.1cm 0 0},clip]{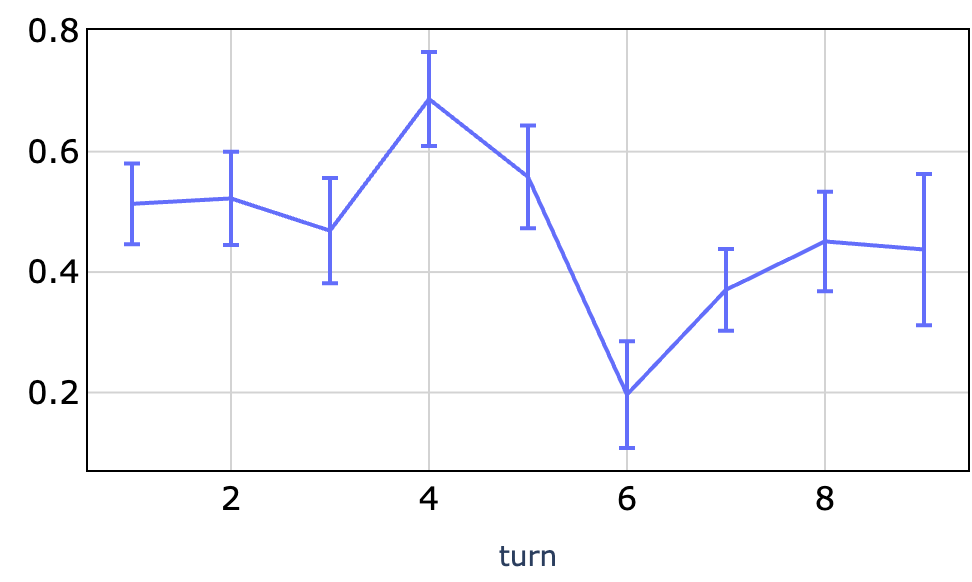}
        \caption{Llama-3.1-70B: Frequency of correct answers for turns 1-9 after seeing a new rule .}
        \label{fig:reverse-digit}
    \end{subfigure}
    \caption{Model performance on the WCST task shows that the models have difficulty inferring and maintaining an underlying rule. The pattern suggests a \textit{failure to maintain set}: it seems to infer a rule with a slight rise in accuracy, but the accuracy sharply drops, indicating ``forgetting'' of the current rule.}
    \label{fig:WCST-res}
\end{figure}


Results (Table~\ref{tab:wcst_performance}) show that \textit{all LLMs perform well below human accuracy}, which is typically 70–80\% \citep{barcelo1997wisconsin, grant1948behavioral, milner1963effects}. 
Unlike humans, models fail to respond well to additional exposure of a given rule, suggesting difficulty in both inferring the underlying rule and maintaining it based on the feedback received within the dialogue (Fig.~\ref{fig:WCST-res}).\footnote{This pattern may echo \citet{pmlr-v235-coda-forno24a}, who found that LLMs under-weight observations in decision-making tasks.} 
For context, in humans, WCST deficits of this magnitude would be consistent with mild cognitive impairment \citep{hammers2016diagnostic, silva2007typical}. While LLMs can perform well on highly advanced benchmarks, their failure to adapt to the WCST may indicate potential weaknesses in cognitive flexibility. The WCST may also hit on the same weaknesses LLMs have in completing the $n$-back task: both tasks are multi-turn and require constant updating of information.


\begin{table}[tbp]
\centering
\small
\begin{tabular}{clccc}
\toprule
& & \shortstack{Accuracy \\ ($\uparrow$)} & \shortstack{Preservation \\ Error ($\downarrow$)} & \shortstack{Other \\ Error ($\downarrow$)} \\
\midrule
\multirow{2}{*}{Gemma2} 
&9B  & 0.29 & 0.21 & 0.51 \\
&27B & 0.49 & 0.20 & 0.30 \\
\midrule
\multirow{2}{*}{Llama3.1} 
&8B  & 0.53 & 0.32 & 0.15 \\
&70B & 0.50 & 0.26 & 0.24 \\
\midrule
\multirow{2}{*}{Qwen2} 
&7B  & 0.52 & 0.33 & 0.14 \\
&72B & 0.51 & 0.29 & 0.21 \\
\midrule
\multicolumn{2}{r}{\textit{Healthy adults}} 
 & 0.77 & 0.12 & 0.09\\
\bottomrule
\end{tabular}
\caption{Frequencies of correct answers and errors in WCST (due to rounding, all do not sum to exactly 1) across all 3060 trials (102 trials x 30 prompts). Human norms naturally vary slightly across studies; this table shows those reported in \citet{barcelo1997wisconsin}. }
\label{tab:wcst_performance}
\end{table}


\subsubsection*{LLMs flanker task performance is below human baselines}\label{sec:flanker}
Executive functioning depends not only on flexibility but also on attentional control, the ability to prioritize task-relevant information while suppressing distractors. This faculty is tightly coupled to working memory capacity \citep{kane2003working, pratt2011effects}, both relying on a central “attentional spotlight.”

To test LLMs on attentional control, we use the classic Eriksen flanker task \citep{eriksen1974effects}. In this task, participants respond to the central letter of a string: e.g., hit one button if the center letter is X or C, and another button if the center letter is B or V. Participants must ignore the ``flanker'' letters on either side of the center. Two types of sequences are shown:
\begin{itemize}[noitemsep,topsep=0pt]
    \item \textit{Congruent trials}(e.g., `X X C X X') have flankers with the same response as the target (`X' and `C').
    \item \textit{Incongruent trials} (e.g., `B B C B B') place the flankers `B' in conflict with target `C,' requiring good attentional control to ignore the flanker.
\end{itemize}
The key metric in this task is reaction time, which typically varies from 300-500ms. Participants have longer reaction times for incongruent strings because inhibiting the flankers requires additional cognitive resources, which slows the response.

\begin{figure}[t]
    \centering
    \begin{subfigure}[b]{0.48\linewidth}
        \includegraphics[trim={0 0 120 0}, 
                     clip,width=\linewidth]{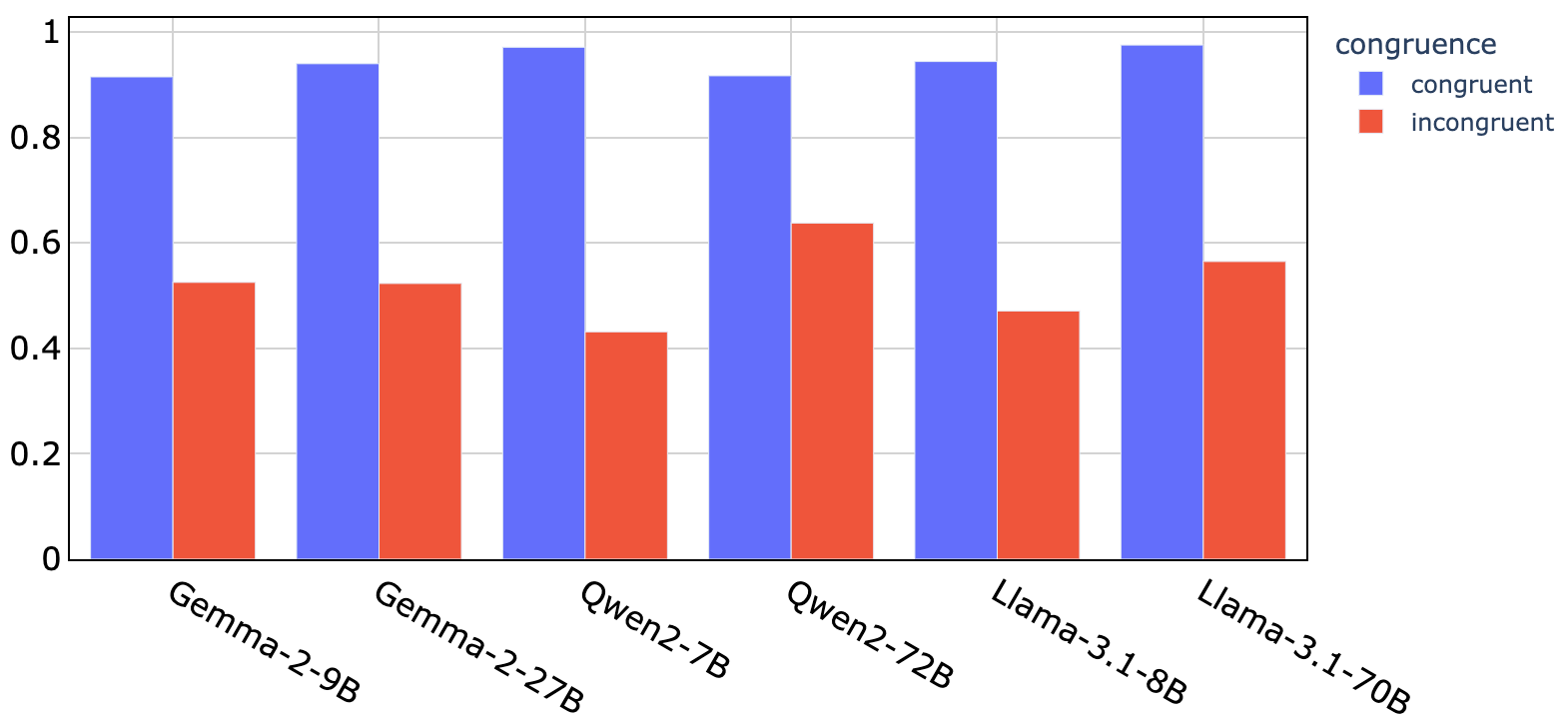}
        \caption{Frequencies of correct responses on the flanker task. All models do substantially worse in the incongruent condition. 
        }
        \label{fig:flanker}
    \end{subfigure}
    \hfill
    \begin{subfigure}[b]{0.48\linewidth}
        \includegraphics[trim={0 0 120 0}, 
                     clip,width=\linewidth]{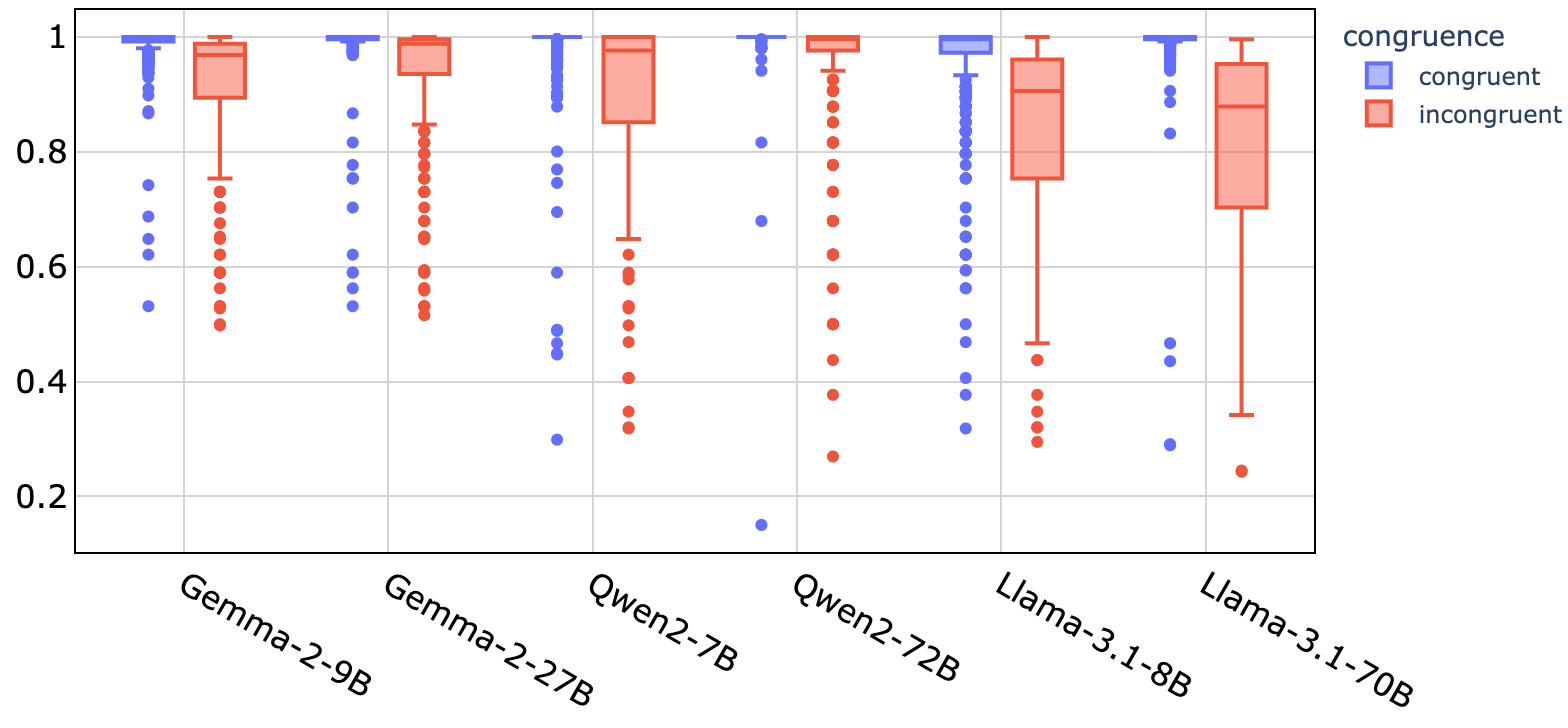}
        \caption{For correct answers, confidence (probability of the correct answer) decreases in the incongruent condition, indicating interference from the flanking letters.}
        \label{fig:flanker-prob}
    \end{subfigure}
    \caption{Flanker task: (a) average accuracy and (b) probabilities of correct answers. All models tested perform worse in the incongruent condition.}
    \label{fig:flanker-digit}
\end{figure}

Because we cannot measure reaction times in LLMs, we instead measure accuracy and the probabilities assigned to correct answer. We find that model perform substantially worse in incongruent trials, with accuracies of 40-60\%, compared to human performance of \textbf{93\% to 96\%} \cite{eriksen1974effects, yantis1990locus}.
Even when responses are correct, model confidence drops in incongruent cases (Figure \ref{fig:flanker-digit}), indicating a vulnerability to the distracting flankers.\footnote{We verify that tokenization of the letters never groups them such that two letters share the same token.}


Together, the WCST and flanker task results suggest that \textit{LLMs may lack the executive mechanisms needed to strategically utilize their large working memory capacity}.
While they can access and reproduce long sequences, they \textit{struggle with selecting relevant information, suppressing irrelevant inputs, and updating information flexibly}.

\begin{figure}[h!]
    \centering
    \begin{subfigure}[b]{\linewidth}
        \centering
        \includegraphics[width=\linewidth]{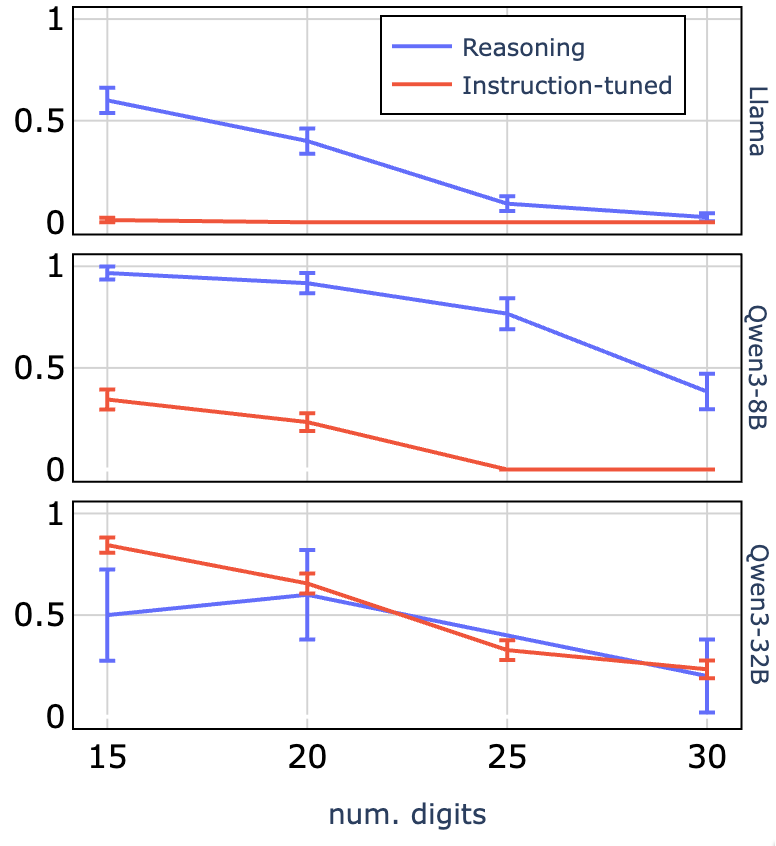}
        \caption{
        Backward digit span accuracy across sequence lengths for reasoning vs. non-reasoning models. Reasoning aids weaker models (Llama-8B, Qwen-8B), but does not achieve perfect accuracy.
        }
        \label{fig:reasoning-bds}
    \end{subfigure}
    \hfill
    \begin{subfigure}[b]{\linewidth}
        \centering
        \includegraphics[width=\linewidth]{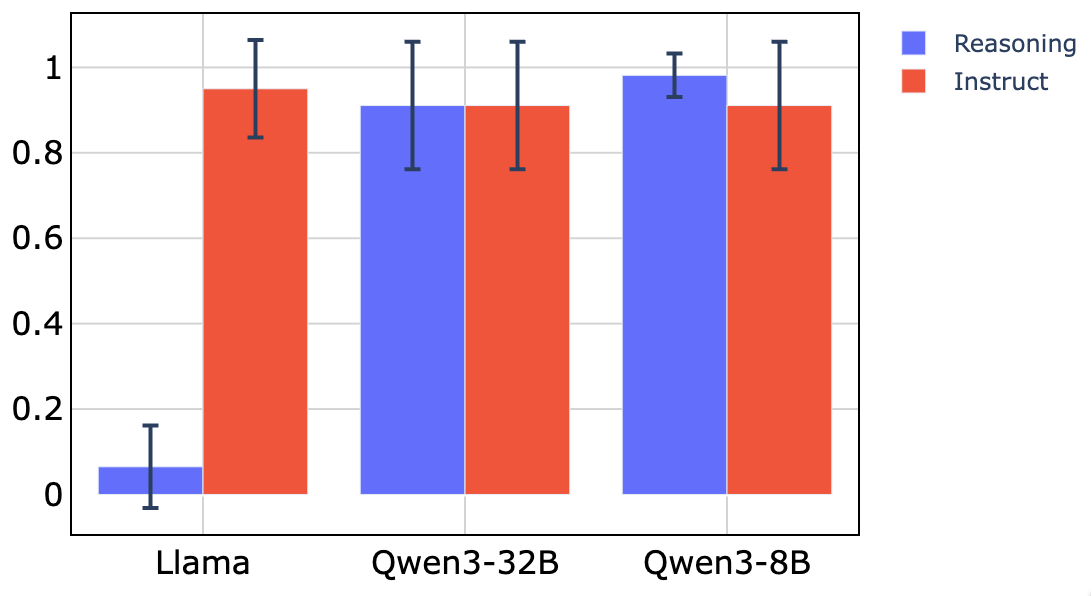}
        \caption{
        Reading span accuracy. Reasoning variants do not outperform instruction-tuned.
        }
        \label{fig:reasoning-ospan}
    \end{subfigure}
    \caption{Reasoning does not reliably improve working memory capacity. We find that the majority of errors are due to models failing to converge during their thought string or hallucinating during the thought string.}
    \label{fig:reasoning-wmc}
\end{figure}

\section{Capabilities of Reasoning models}\label{sec:reasoning}
We next ask whether reasoning-enabled variants of LLMs show advantages of working memory and executive function.
The rationale is that the intermediate ``thought strings'' produced by reasoning models might act as a pseudo central executive, supporting memory maintenance and executive attention or planning.
Specifically, we evaluate Qwen3-8B and Qwen3-32B, with and without reasoning enabled, and compare Llama-3.1-8B (instruction-tuned) variant against its reasoning variant distilled from DeepSeekR1.

\subsection{Reasoning does not consistently improve working memory capacity}
Working memory capacity was tested using the backward digit span and reading span tasks (Sections~\ref{sec:wmc}, \ref{sec:wmc_complex}). Across both tasks, reasoning variants did not yield consistent improvements (Figure~\ref{fig:reasoning-wmc}). While reasoning sometimes boosted weaker baselines (e.g., Llama-8B, Qwen-8B on backward digit span), accuracy remained well below ceiling and failed to generalize. 
Inspection of thought strings suggests a tendency toward \textit{excessive deliberation}: models often looped through near-identical reasoning steps without converging, despite the very simple and straightforward natures of the tasks \cite{chennot}; for examples see \S\ref{sec:thoughts}. 
When reasoning models do markedly worse, we find it is because thought strings fail to converge, possibly due to the tasks being out-of-domain.

\subsection{Reasoning improves attentional control on flanker task (at high computational cost)}
In contrast, reasoning models show clear benefits on flanker task accuracy (Table~\ref{tab:reasoning_flanker}). They outperform their non-reasoning counterparts on incongruent trials, though \textit{at high computational cost}: models often generate thousands of tokens to solve a task humans complete in 300-500ms. Notably, reasoning sequences were equally long for congruent and incongruent conditions, suggesting that models do not capitalize on the simplicity of congruent cases (all letters require the same response, so it does not really matter which letter is in the middle). Instead, they adapt thought heuristics likely learned during training, such as: \textit{``Maybe the middle letter isn’t the exact center, but the third in the sequence.''} These strategies appear inefficient.

\begin{table}[t]
\centering
\begin{tabular}{lcc}
\toprule
\textbf{Model} & \textbf{Congruent} & \textbf{Incongruent} \\
\midrule
Llama-3.1-8B            & 0.944 & 0.471 \\
\quad +Reasoning & 0.977 & \textbf{0.910} \\
\midrule
Qwen2-32B               & 0.917 & 0.638 \\
\quad +Reasoning & 0.994 & \textbf{0.985} \\
\bottomrule
\end{tabular}
\caption{Flanker accuracies (simple attentional control) on the congruent vs. incongruent conditions. Adding reasoning allows models to achieve much higher accuracies in this task. Thought strings are several hundred tokens long, despite the straightforward and simple task.}
\label{tab:reasoning_flanker}
\end{table}

\subsection{Reasoning does not improve strategic planning on WCST}
We also tested reasoning models on the Wisconsin Card Sorting Task (Section~\ref{sec:cognitive}). Instruction-tuned LLMs performed substantially below healthy adult baselines, failing to adapt to feedback when inferring the rule. Because strategic planning could, in principle, benefit from stepwise reasoning, we added explicit strategy hints to the prompt (guidance that is not available to human participants):

\begin{tcolorbox}[colback=blue!5!white,colframe=blue!75!black,title=Hints added to WCST prompts]
\small
  \textbf{To instructions:} (...) There is a simple strategy you can follow: 
  
  (1) if there is a previous answer, check the feedback. 
  
  (1a) If the feedback says correct, just apply the same rule. 
  
  (1b) If the feedback says incorrect, randomly pick a *different rule* to try. 
  
  (2) If there is no previous answer, randomly try a rule. 

  \tcblower
  \small
    \textbf{To each messages:} Remember that the underlying rule can change, so the most recent feedback (from the last 1-2 turns) is the most important to consider -- you should ignore all but the most recent messages.
\end{tcolorbox}

However, even with this explicit scaffolding, reasoning models failed to outperform their non-reasoning counterparts (Table~\ref{tab:wcst_performance}). Thought strings often failed to terminate under our 30,000 token limit (a generous threshold, given that humans make their choice in seconds). Analysis of thought strings revealed frequent hallucinations, irrelevant elaborations, and fabricated stimuli, rather than systematic strategic adjustment. The thought strings also frequently disregard the instructions on task strategy; for instance Qwen3-8B generates: ``Wait, the previous rule was count, and the feedback was correct. But maybe the rule changed?'' which directly contradicts the instructions.

\section{Discussion}
Our experiments consistently suggest that LLMs have a larger working memory capacity than humans do. 
This is not wholly unsurprising, given the propensity the transformers architecture has for copying in-context text. However, we critically find that LLM memory tends to get weaker when manipulation of in-context information is required (even a simple reversal, as in the backward digit span), when updating is required (e.g., $n$-back task), or when simultaneous processing is required (i.e., operation span or reading span).

Our results further suggest that attentional control -- or selecting relevant information to focus on while ignoring distractors -- is another vulnerability for LLMs. Notably, reasoning models were able to successfully mitigated this in the flanker task. 

The \textit{updating} difficulty may also interfere with LLM ability to adapt to the WCST. While humans have a relatively easy time inferring a simple sorting rule based on feedback, no language model approached human-like performance on this task. If state updating is too difficult, then LLMs will not be able to successfully update their beliefs after receiving feedback, leading to low accuracy scores like the ones we observe.

While reasoning models can possibly supplement these difficulties by, e.g., generating thought strings that repeat information in a strategic, goal-oriented way to aid retrieval, we do not find consistent evidence of such behavior. Instead, the reasoning models tested could augment performance on simpler, one-turn tasks (e.g., Flanker, backward digit span) and did not help on the most complex, longest task (WCST). 

It is possible that reasoning models are more suited to certain cognitive functions, explaining the asymmetric success we observe. However, it should be noted that reasoning models are trained on thought strings for vastly different types of problem solving, and their learned thought heuristics (e.g., ``Wait,'' or ``let me double check'') do not appear to serve them well on these simple cognitive tasks. With more cognitively-focused training and more efficient thought strings, reasoning models could effectively augment executive functioning in LLMs, unlocking more intelligent models.

\section{Conclusion}
We apply a suite of working memory tasks to LLMs, finding that they consistently exceed human performance. 
In contrast, we find that the LLMs tested do poorly on tasks of executive functioning. 
Manipulation and updating of information are identified as key cognitive difficulties for LLMs, and some difficulties with attentional control are also observed. 
While reasoning models show promise in compensating for these cognitive deficits, they are highly inefficient and not well-trained to do so.

\section*{Limitations}
We report only behavioral results that cannot be used to interpret what a model encodes in its internal representations. We believe these behavioral results are revealing, but as they only capture the generation process, they do not shed light on the models' internal computations or knowledge. 

We focus on a collection of six instruction-tuned and three reasoning models; it is possible that other language models will not follow the trends found here, although we have made efforts to collect a reasonable sample of models.

We make efforts to test a range of prompts and pilot test prompting strategies, but we are unable to definitively determine that our prompts are representing the full range of possible model responses to these tasks.

\section*{Ethics Statement}
In trying to address the working memory limitations discussed in this paper, it may be tempting to turn to scaling up LLMs, whether via additional attention heads, wider context windows, generating more reasoning tokens, ensembling multiple language models, or some combination thereof.

We instead advocate for research toward more efficient models.
There are very pressing environmental concerns associated with generative language models, and further scaling reasoning tokens should not be done without regard to the sustainability of the models. 
This is particularly relevant as this paper addresses working memory, a key feature for models in industrial, user-facing applications.


\bibliography{custom}

\appendix
\section{Appendix}
\subsection{WCST for Vision Language Models}\label{sec:wcst-vlmodels}
We create both a vision-text and text-only version of the WCST, but as the vision language models tested on the visual WCST perform substantially worse than text models (Table~\ref{tab:vl_wcst}), we focus our analysis on text-only versions of the task.

\begin{figure}
    \centering
    \includegraphics[width=\linewidth]{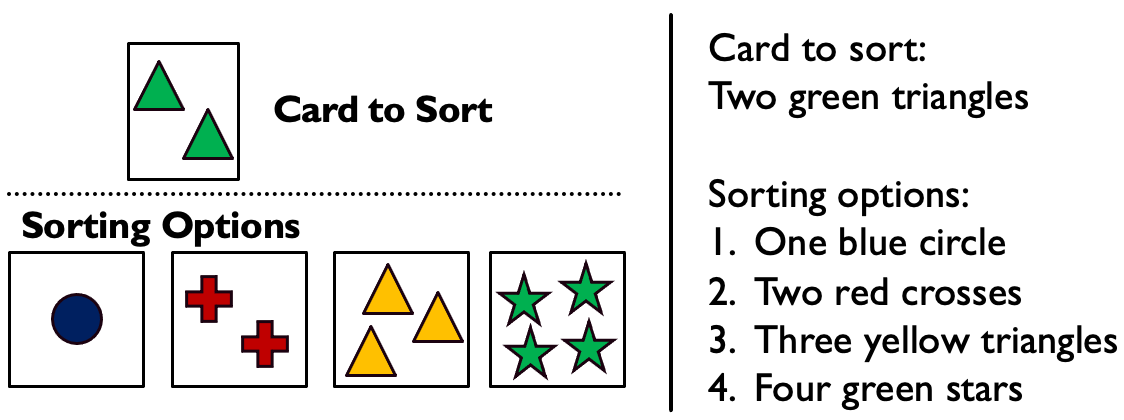}
    \caption{We create a text version of the WCST for LLMs (right). We use some VLMs to test a visual version of the WCST (left).
    }
    \label{fig:wcst}
\end{figure}

The vision version of the task displays the options and the card to sort as an image, as shown in Fig.~\ref{fig:wcst}. The instructions otherwise mirror those in the text-based WCST.

\begin{table}[h]
\small
\centering
\begin{tabular}{lccc}
\toprule
\textbf{Model} & \textbf{Accuracy} & \textbf{Err (P)} & \textbf{Err (O)} \\ \midrule
Qwen2.5-VL-32B      & 0.17 & 0.33 & 0.50 \\
Qwen2.5-VL-7B       & 0.11 & 0.40 & 0.49 \\ \midrule
\textit{Qwen2-7B (text-only)}  & 0.44 & 0.50 & 0.06 \\
\bottomrule
\end{tabular}
\caption{Qwen2.5-VL model performance on visual-text WCST fall below Qwen2 text-only WCST performance. Err(P) denotes preservation error rate, while Err(O) denotes other error rate.}
\label{tab:vl_wcst}
\end{table}

\subsection{Overthinking in Models}\label{sec:thoughts}
We observe several detrimental thought patterns in reasoning models. Below are selected examples from Qwen3-8B.

One is incoherent reiterations of the situation, e.g., ``Now, the third letter is being presented. But the user hasn't given it yet,'' or ``Some words are repetitive and nonsensical, but these are the ones that can be clearly recalled,'' from the reading and operation span tasks.

Sometimes we see muddled logic, e.g.: ``The last few interactions included words like `problem,' `speed,' `ice,' and the current one. So, the words to recall would be `ice' and any others mentioned in the current round. But since the user hasn't provided more than one word in this round, the answer should be ``ice.''''

Thought strings will sometimes veer from the task entirely, circling without making progress, e.g.:
``I think I'm getting the hang of it.</br>
So,</br>
I need to make sure that I don't make mistakes.</br>
But I'm still learning.</br>
I think I need to be careful.</br>''

We also note that learned reasoning heuristics (``Wait'' or ``let me double check'') are not very useful in these tasks, further hindering reasoning models. For instance, consider Qwen3-8B's : ``Alright, let's break it down. When I receive a new letter, I need to look back three positions to see if it matches. If it does, I write 'm'; if not, I write '-'. Simple enough, but I have to be careful not to make a mistake with the positions. Maybe I should write them down as I go along to keep track.

Wait, what if the sequence is longer? I mean, how do I handle it if the sequence is, say, 10 letters long? I need to make sure I don't miss any or miscount the positions. Maybe I can keep a running tally or use some kind of notation to mark the positions as I go. That way, when I get to the fourth letter, I can check the first one easily.

Wait, but what if the sequence has letters that repeat or have patterns?''

``Lastly, I should remind myself to stick strictly to the rules. Only 'm' and '-' are allowed, and no explanations. So, I have to make sure my responses are clear and follow the instructions precisely. That way, I can ensure my task is done correctly and efficiently.''

\subsection{Prompt Variations}\label{sec:prompt_spread}
\begin{figure}
    \centering
    \begin{subfigure}[b]{0.255\linewidth}
        \includegraphics[width=\linewidth,trim={1cm 2cm 2cm 3cm},clip]{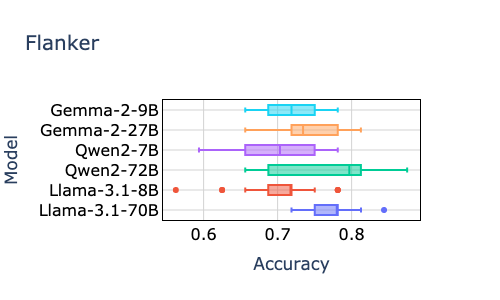}
        \caption{Flanker}
    \end{subfigure}
    \begin{subfigure}[b]{0.175\linewidth}
        \includegraphics[width=\linewidth,trim={5.6cm 2cm 2cm 3cm},clip]{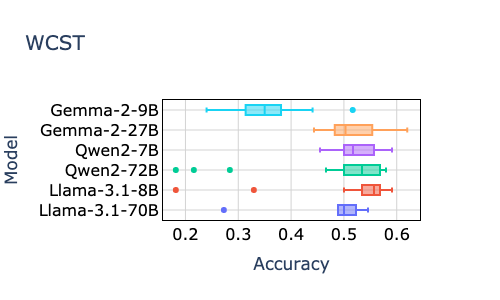}
        \caption{WCST}
    \end{subfigure}
    \begin{subfigure}[b]{0.175\linewidth}
        \includegraphics[width=\linewidth,trim={5.6cm 2cm 2cm 3cm},clip]{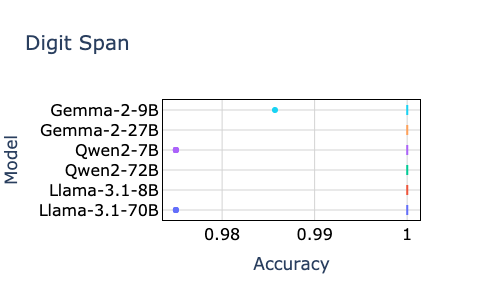}
        \caption{Digit Span}
    \end{subfigure}
    \begin{subfigure}[b]{0.175\linewidth}
        \includegraphics[width=\linewidth,trim={5.6cm 2cm 2cm 3cm},clip]{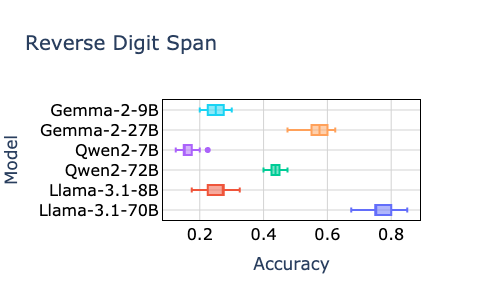}
        \caption{Backward DS}
    \end{subfigure}
    \caption{For each prompt variation, we compute each model's average accuracy on several tasks. These box plots display the range in resulting model accuracies over the prompts. For challenging tasks like Flanker and WCST, the range accuracies are comparatively large.}
    \label{fig:accuracy-spread}
\end{figure}

We assess the extent to which the prompt variations affect the spread of model responses and find that outside of very easy tasks, the variations capture a range of model accuracies (Fig.~\ref{fig:accuracy-spread}).

\subsection{Estimation of human baselines}\label{sec:human-baseline}
It is difficult to precisely estimate human baselines due to the sheer number of studies applying these experimental tasks across a variety of human populations. 

We estimate human baseline accuracy for the backward digit span of length 7 by referring to the means and standard deviations presented in \cite{choi2013normative}. Ideally we would use the Weschler norms, but these are proprietary and we do not have access. Given the backward digit span norms $\mu = 5.4, \sigma=1.5$, we estimate that about 14\% of people could have a backward digit span of 7. For the forward digit span of length 7, multiple sources report means of 7 \citep{banken1985clinical, monaco2013forward}, so we estimate the accuracy at $50\%$.

We use the reported mean human WCST errors in \cite{barcelo1997wisconsin} to estimate human WCST accuracy. The reported mean number of errors is 58.9 across 252 trials, so we estimate the baseline accuracy to be about 77\%.

To estimate n-back accuracy we use Experiment 2 in \citet{jaeggi2010concurrent}.

We use the reported incongruent error rate in the Flanker task from \cite{yantis1990locus}, 4.7\%, to estimate the baseline human accuracy for incongruent stimuli to be about 95\%.

We estimated reading span and operation span accuracy through reported means from Experiment 1 in \cite{broadway2010validating}.

\subsection{Additional Results}

\subsubsection{Forward Digit Span}
The mean accuracies for forward digit span across all models can be found in Table~\ref{tab:forward-digit}
\begin{table*}[ht]
\centering
\small
\begin{tabular}{|l|c|c|c|c|c|c|}
\hline
\textbf{Length} & \textbf{Gemma-2-27B} & \textbf{Gemma-2-9B} & \textbf{Llama-3.1-70B} & \textbf{Llama-3.1-8B} & \textbf{Qwen2-72B} & \textbf{Qwen2-7B} \\
\hline
7   & 1.00 & 1.00 & 0.99 & 1.00 & 1.00 & 1.00 \\
20  & 1.00 & 1.00 & 1.00 & 1.00 & 1.00 & 0.98 \\
30  & 1.00 & 1.00 & 1.00 & 1.00 & 1.00 & 1.00 \\
50  & 1.00 & 0.99 & 0.99 & 1.00 & 1.00 & 1.00 \\
\hline
\end{tabular}
\caption{Mean accuracies on forward digit span across all prompts.}\label{tab:forward-digit}
\end{table*}
.

\subsubsection{Backward Digit Span}
The mean accuracies for forward digit span across all models can be found in Figure~\ref{fig:reverse-digit}.
\begin{figure}
    \centering
        \includegraphics[width=\linewidth,trim={0 0.4cm 0 0},clip]{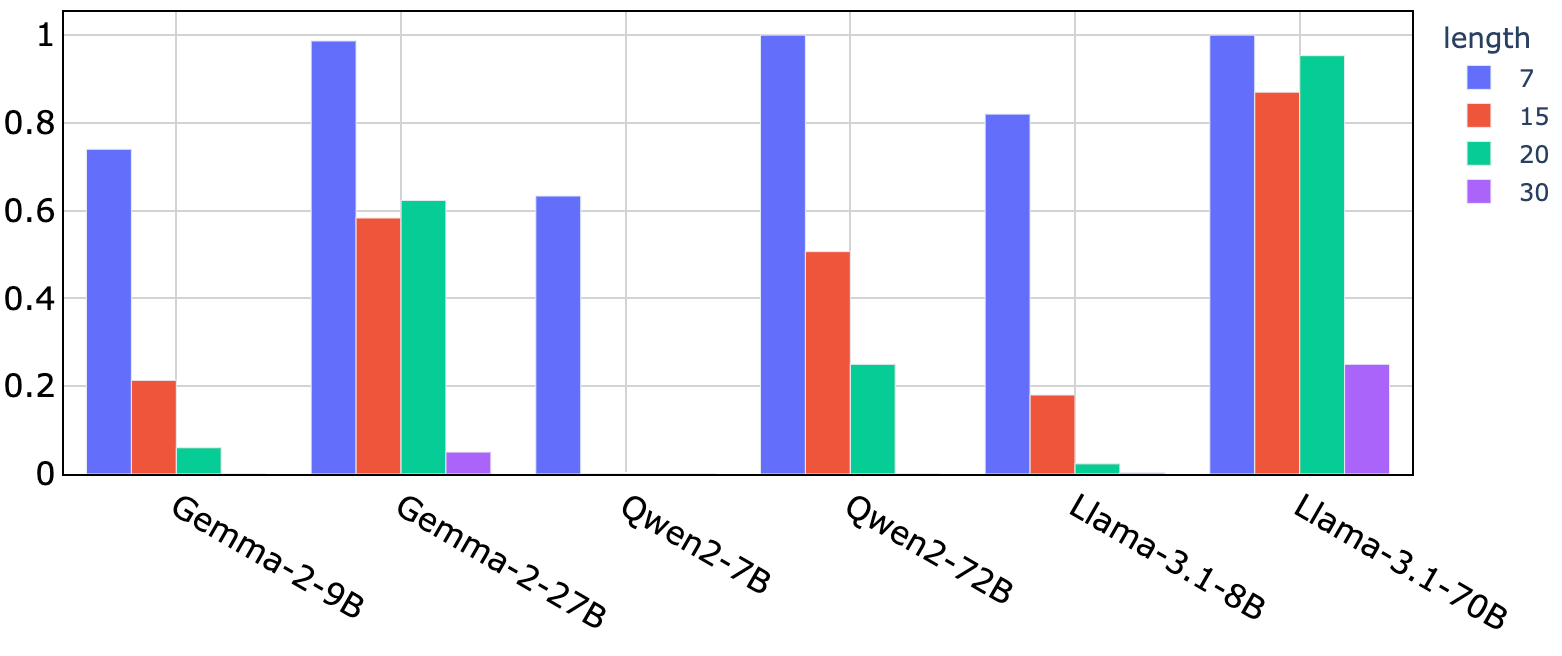}
        \caption{Mean model accuracy, by digit sequence length, in the backward digit span task across all trials ($n = 2100$, 70 stimuli x 30 prompts).
        }
        \label{fig:reverse-digit}
\end{figure}

\subsubsection{Wisconsin Card Sorting Task}\label{sec:WCST-appendix}
\begin{figure}
    \centering
    \begin{subfigure}[b]{0.255\linewidth}
        \includegraphics[width=\linewidth,trim={1cm 2cm 2cm 3cm},clip]{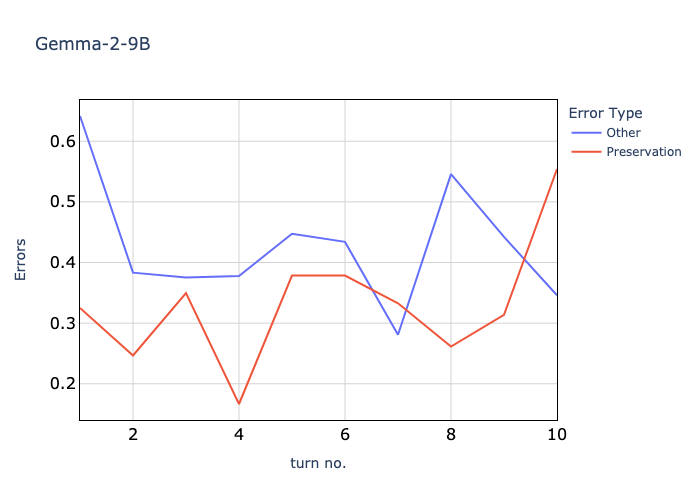}
        \caption{Gemma2 (9B)}
    \end{subfigure}
    \begin{subfigure}[b]{0.255\linewidth}
        \includegraphics[width=\linewidth,trim={1cm 2cm 2cm 3cm},clip]{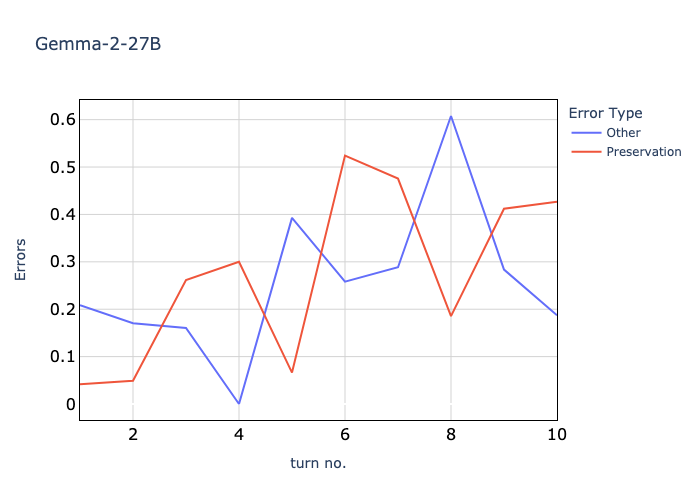}
        \caption{Gemma2 (27B)}
    \end{subfigure}
    \begin{subfigure}[b]{0.255\linewidth}
        \includegraphics[width=\linewidth,trim={1cm 2cm 2cm 3cm},clip]{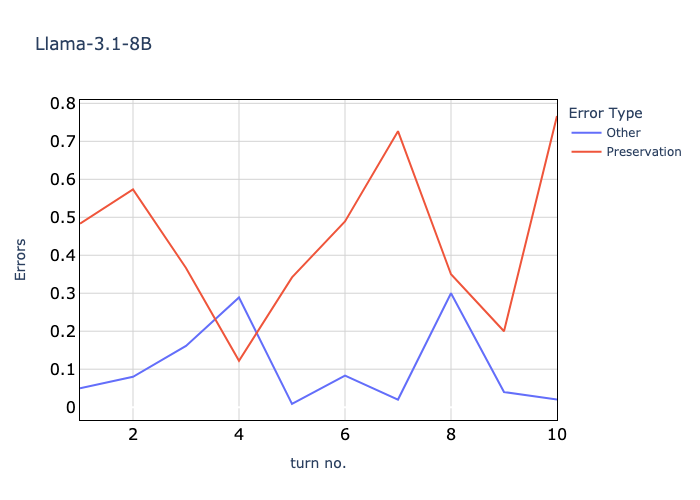}
        \caption{Llama3 (8B)}
    \end{subfigure}
    \begin{subfigure}[b]{0.255\linewidth}
        \includegraphics[width=\linewidth,trim={1cm 2cm 2cm 3cm},clip]{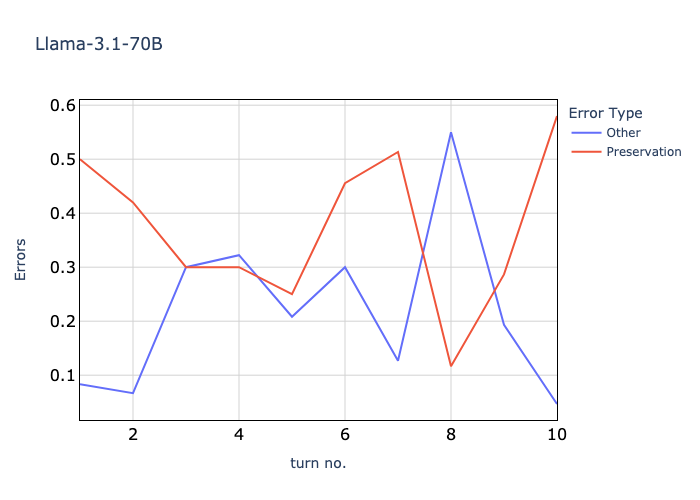}
        \caption{Llama3 (70B)}
    \end{subfigure}
    \begin{subfigure}[b]{0.255\linewidth}
        \includegraphics[width=\linewidth,trim={1cm 2cm 2cm 3cm},clip]{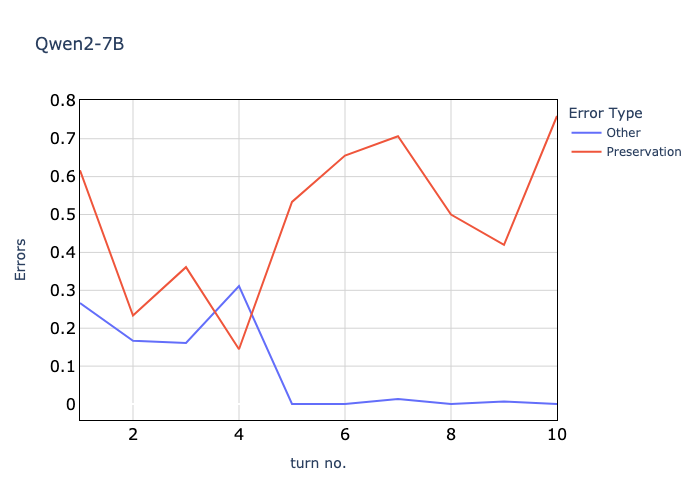}
        \caption{Qwen2 (7B)}
    \end{subfigure}
    \begin{subfigure}[b]{0.255\linewidth}
        \includegraphics[width=\linewidth,trim={1cm 2cm 2cm 3cm},clip]{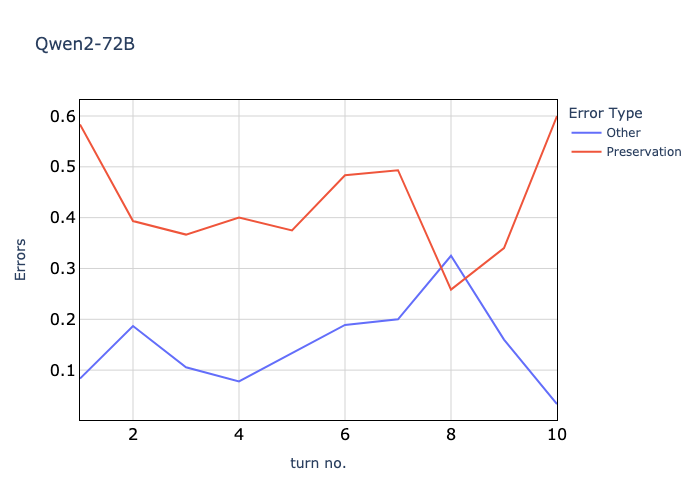}
        \caption{Qwen2 (72B)}
    \end{subfigure}
    \caption{We take the average error rate for preservation (red) and other (blue) errors across the first 10 rounds after a new rule is introduced. We find no correlation between any type of model error and the number of rounds exposed to a new rule.}
    \label{fig:wcst-errs}
\end{figure}

The average error rates over the first ten turns after a rule is introduced are shown in Figure~\ref{fig:wcst-errs} for each model.

\begin{figure}
    \begin{subfigure}[b]{0.45\linewidth}
        \includegraphics[width=\linewidth,trim={1cm 2cm 2cm 1cm},clip]{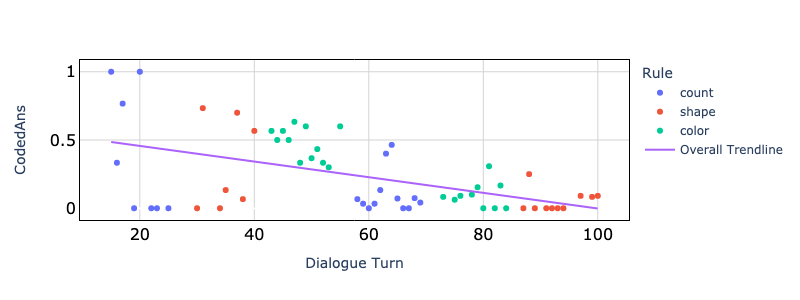}
        \caption{Gemma 2 (9B)}
    \end{subfigure}
    \hfill
    \begin{subfigure}[b]{0.45\linewidth}
        \includegraphics[width=\linewidth,trim={1cm 2cm 2cm 1cm},clip]{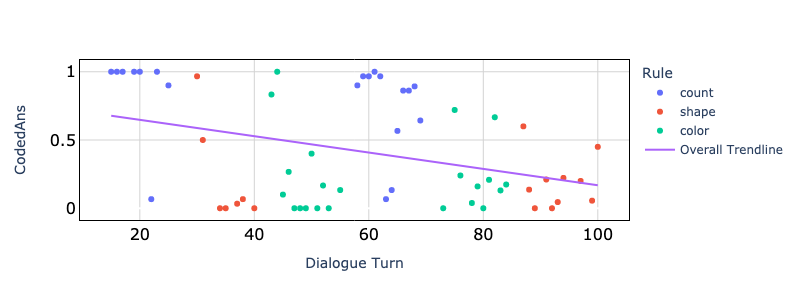}
        \caption{Gemma 2 (27B)}
    \end{subfigure}
    \begin{subfigure}[b]{0.44\linewidth}
        \includegraphics[width=\linewidth,trim={1cm 2cm 2cm 1cm},clip]{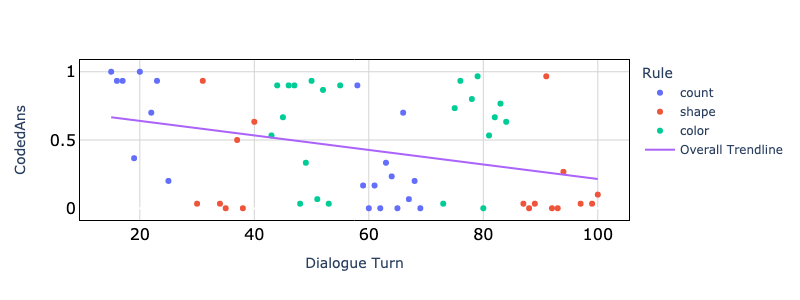}
        \caption{Llama 3.1 (8B)}
    \end{subfigure}
    \hfill
    \begin{subfigure}[b]{0.45\linewidth}
        \includegraphics[width=\linewidth,trim={1cm 2cm 2cm 1cm},clip]{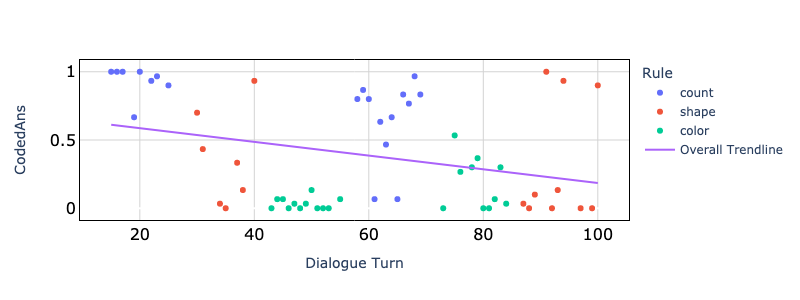}
        \caption{Llama 3.1 (70B)}
    \end{subfigure}
        \begin{subfigure}[b]{0.4\linewidth}
        \includegraphics[width=\linewidth,trim={1cm 2cm 2cm 1cm},clip]{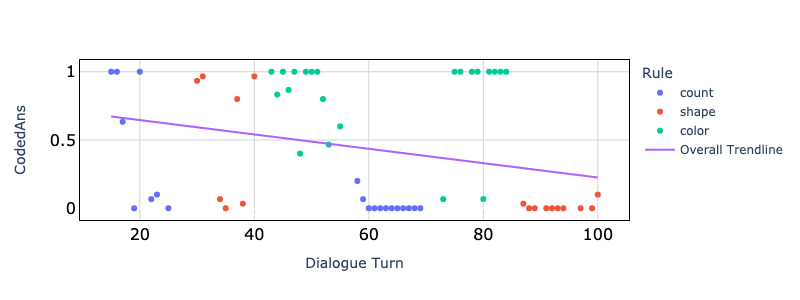}
        \caption{Qwen 2 (7B)}
    \end{subfigure}
    \hfill
            \begin{subfigure}[b]{0.45\linewidth}
        \includegraphics[width=\linewidth,trim={1cm 2cm 2cm 1cm},clip]{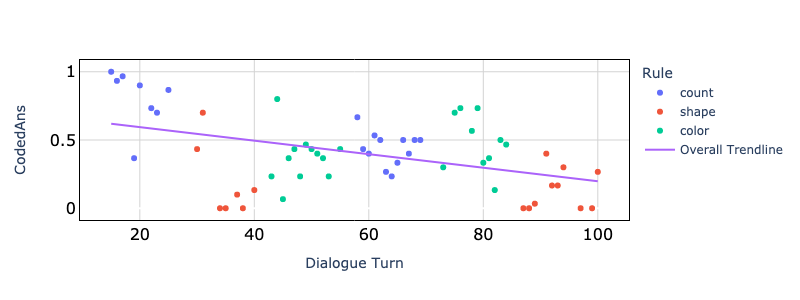}
        \caption{Qwen 2 (72B)}
    \end{subfigure}
    \caption{Trends for model accuracy over the course of the WCST dialogue.}
    \label{fig:wcst-trends}
\end{figure}

We also investigate model performance over all dialogue turns. Correlation results are in Table~\ref{tab:wcst-corr}, and the remaining model plots are in Figure~\ref{fig:wcst-trends}.

\begin{figure}
    \centering
    \includegraphics[width=0.8\linewidth]{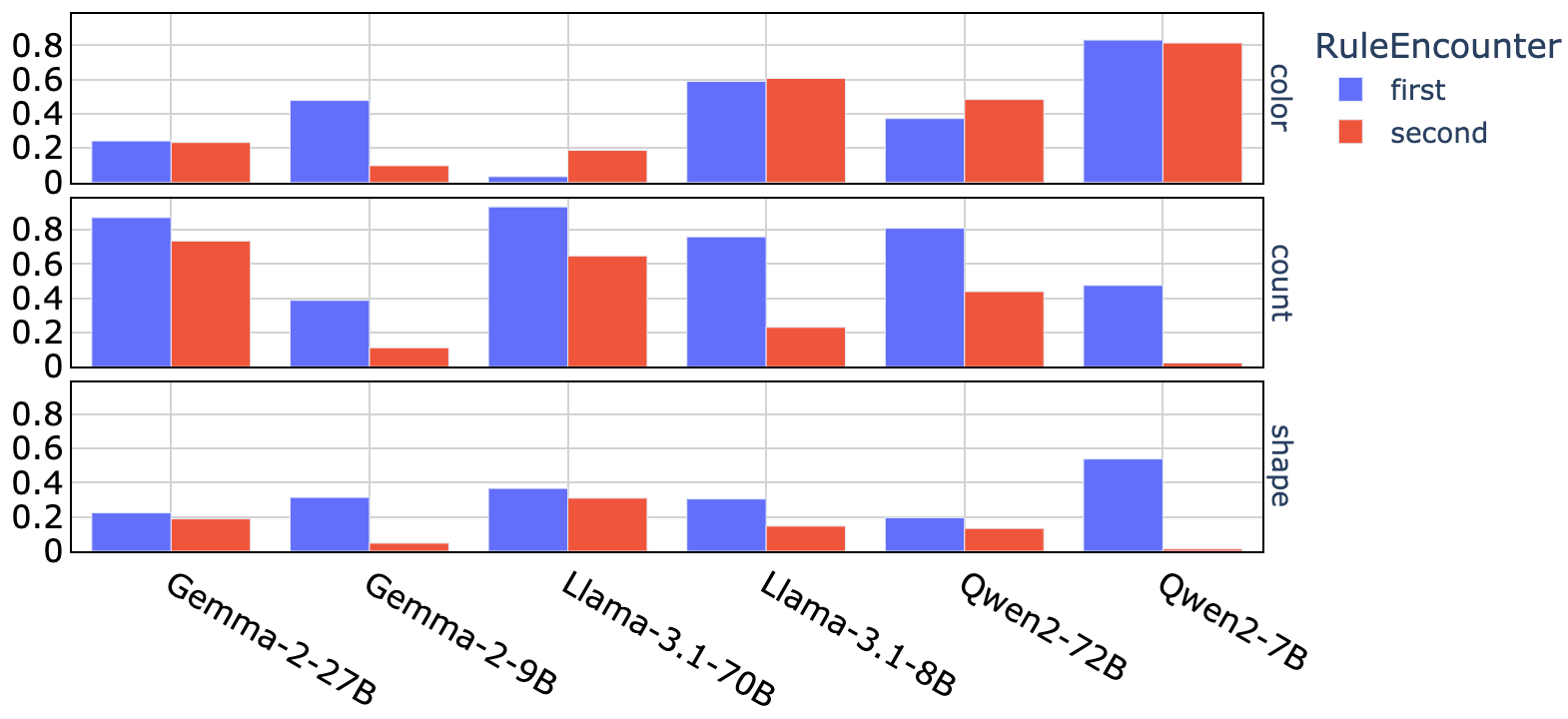}
    \caption{Models tend to have lower accuracy the second time they are exposed to the count and shape sorting rules (second and third rows).}
    \label{fig:wcst-second}
\end{figure}

Finally, decreases in accuracy between first and second exposure to a sorting rule can be seen in Figure~\ref{fig:wcst-second}

\begin{table*}[h!]
\centering
\small
\begin{tabular}{|l|c|c|c|c|c|c|}
\hline
 & \textbf{Gemma-27B} & \textbf{Gemma-9B} & \textbf{Llama-70B} & \textbf{Llama-8B} & \textbf{Qwen-72B} & \textbf{Qwen-7B} \\
\hline
Pearson's $r$ & -0.36* & -0.51* & -0.31* & -0.33* & -0.45* & -0.28* \\
\hline
\end{tabular}
\caption{Pearson correlation results for model accuracy and dialogue length in the WCST. Significant correlations ($p < 0.05$) are marked with *. We find all model accuracy decreases as the dialogue goes on.}\label{tab:wcst-corr}
\end{table*}

\section{Licenses of Models}
All licenses are fair use for this work. 
\begin{enumerate}
    \item LLaMA-3.1 has a Llama 3.1 Community License.
    \item RoBERTa has an MIT license.
    \item The nvidia/HelpSteer2 dataset has a Creative Commons Attribution 4.0 license.
    \item The Anthropic/hh-rlhf dataset has an MIT license.
    \item The OpenAssistant/oasst1 dataset has an Apache 2.0 license.
\end{enumerate}

\subsection{Prompt Specifications}\label{sec:prompts}
Prompts from the reading span task:
\begin{promptbox}
    The following task consists of two parts. First, you will be presented with sentences, which you are supposed to read  carefully. These sentences can make sense (e.g. More and more women want to have a career.) or not (e.g. I stopped by the gas station to refuel on apple juice.). Your first task is to judge the content of each sentence. You will then respond with your judgment (True = yes it makes sense, False = no it does not). After this decision you will see a word on the screen, which you are also supposed to read carefully. Your second task is to memorize this word in order to recall it later on. After two to six sentences (which you are supposed to judge) and words (which you are supposed to memorize), you will see three question marks (???). Whenever this happens, please respond by recalling all words you still remember since the last ???. Often it’s easier to recall them in the order in which they were presented, but this isn’t really necessary. If you only remember a couple, but not all words, it’s okay to only name the ones that you do remember. After you have named all these words, the next round will begin. Your answers will all be recorded. We will now begin.
\end{promptbox}

Prompts from the backward digit span:
\begin{promptbox}
    You will be presented with a list of digits. Repeat the digits back, in reverse order. Your response should contain *only* the reversed digits. Below are some examples of number sequences, followed by the reverse sequence:Digits: 1,2,3,4,5    5,4,3,2,1        Digits: 6,1,2,3,1,8,7    7,8,1,3,2,1,6     This concludes the examples. Below is a list of digits; respond by repeating these digits back, in reverse order. Your response should contain *only* the reversed digits.
\end{promptbox}

Prompts from the flanker task:
\begin{promptbox}
    you'll be presented with 5 to 11 letters. Focus on the letter in the center. When encountering an X or a C, reply with \"A.\" If a V or a B appears, reply with \"L.\" Your response should solely consist of \"A\" or \"L,\" excluding any other letters. For example, if you see the letters \"BBCBB\"     your response should be exactly \"A\". If you see the letters \"XXXXCXXXX\" your response should be exactly \"A\". Only respond with a single letter. This is the end of the instructions; the letters are below:
\end{promptbox}

Below, we present the data formats and prompt templates used for the WCST to illustrate the prompt perturbation process.

\begin{promptbox}[Original Prompt Template - WCST]
You will see an item, and you will have to match it to one of four option. Your task is to figure out the classification rule to sort the item into one of the four options.

Respond with *only the option number* you would like to sort the card into. After you respond, you will get feedback about your response. You will have to try a different classification rule if the feedback says you were wrong. This is an example in which the underlying rule is color:

[EXAMPLE START]

Item: two red circles
Options: 1. one red circle, 2. two blue crosses, 3. three yellow stars, 4. four green triangles

---

Answer: 1

---

Feedback: Correct

[EXAMPLE END]

Here's an example in which the underlying classification rule is *count*:

[EXAMPLE START]
Item: two yellow triangles
Options: 1. one red circle, 2. two blue crosses, 3. three yellow stars, 4. four green triangles

---

Answer: 4

---

Feedback: Correct
[EXAMPLE END]

Here's an example in which the underlying classification rule is *shape*:

[EXAMPLE START]

Item: two green circles
Options: 1. one red circle, 2. two blue crosses, 3. three yellow stars, 4. four green triangles

---

Answer: 1

---

Feedback: Correct

[EXAMPLE END]

Now we will begin.

\verb|<<DATA>>|
\end{promptbox}

\begin{promptbox}[Paraphrased Prompt Template 1 - WCST]
You will be shown an item, and your task is to match it with one of four options. Your objective is to determine the hidden classification rule that assigns the item into one of these four options. The classification rule may be shape, color, or count.

Reply with *only the option number* you believe the item should be matched with, based on the classification rule. After you reply, you will receive feedback regarding your choice. If the feedback says you were incorrect, you will need to attempt a different classification rule. Note that the rule may change at any point; keep using the feedback to figure out the current rule. Here's an example in which the underlying rule is color:

[EXAMPLE START]

Item: two red circles  
Options: 1. one red circle, 2. two blue crosses, 3. three yellow stars, 4. four green triangles

---

Answer: 1

---

Feedback: Correct

[EXAMPLE END]

Here's an example in which the underlying classification rule is *count*:

[EXAMPLE START]  
Item: two yellow triangles  
Options: 1. one red circle, 2. two blue crosses, 3. three yellow stars, 4. four green triangles

---

Answer: 4

---

Feedback: Correct  
[EXAMPLE END]

Here's an example in which the underlying classification rule is *shape*:

[EXAMPLE START]

Item: two green circles  
Options: 1. one red circle, 2. two blue crosses, 3. three yellow stars, 4. four green triangles

---

Answer: 1

---

Feedback: Correct

[EXAMPLE END]

Let's get started.

\verb|<<DATA>>|
\end{promptbox}

\begin{promptbox}[Paraphrased Prompt Template 2 - WCST]
You will observe an object and need to categorize it into one of four options. Your goal is to determine the rule that classifies the object into one of these options.

Reply with *only the option number* where you believe the object belongs. Once you submit your answer, you'll receive feedback on whether your classification was correct. If it was incorrect, you'll need to revise your classification strategy. Below is an example in which the underlying rule is color:

[EXAMPLE START]

Item: two red circles
Options: 1. one red circle, 2. two blue crosses, 3. three yellow stars, 4. four green triangles

---

Answer: 1

---

Feedback: Correct

[EXAMPLE END]

Next is an example in which the underlying classification rule is *count*:

[EXAMPLE START]Item: two yellow triangles
Options: 1. one red circle, 2. two blue crosses, 3. three yellow stars, 4. four green triangles

---

Answer: 4

---

Feedback: Correct
[EXAMPLE END]

Finally, an example in which the underlying classification rule is *shape*:

[EXAMPLE START]

Item: two green circles
Options: 1. one red circle, 2. two blue crosses, 3. three yellow stars, 4. four green triangles

---

Answer: 1

---

Feedback: Correct

[EXAMPLE END]

Now let's start...

\verb|<<DATA>>|
\end{promptbox}

\begin{promptbox}[Data Format 1 (original) - WCST]
\begin{tabbing}
Feedback: \verb@(|FEEDBACK TEXT|)@ \\
\\
Item: \verb@(|CARD TO SORT|)@ \\
\\
Options: \\
1. \= one red circle, \\
2. \= two green triangles, \\
3. \= three blue crosses, \\
4. \= four yellow stars \\
\end{tabbing}
\end{promptbox}

\begin{promptbox}[Data Format 2 (Field: \{\}\textbackslash n Answer: \{\}) - WCST]
\begin{tabbing}
Feedback: \verb@(|FEEDBACK TEXT|)@ \\
\\
Item: \verb@(|CARD TO SORT|)@ \\
\\
Options: \= 1. one red circle, 2. two green triangles, 3. three blue crosses, 4. four yellow stars \\
\end{tabbing}
\end{promptbox}

\begin{promptbox}[Data Format 3 (Field: \{\} \texttt{<sep>} Answer: \{\}) - WCST]
\begin{tabbing}
Feedback: \verb@(|FEEDBACK TEXT|)@ \= \texttt{<sep>} \\
Item: \verb@(|CARD TO SORT|)@ \= \texttt{<sep>} \\
Options: 1. one red circle, 2. two green triangles, 3. three blue crosses, 4. four yellow stars \\
\end{tabbing}
\end{promptbox}

\begin{promptbox}[Data Format 4 (Field - \{\}. Answer - \{\}) - WCST]
\begin{tabbing}
Feedback - \verb@(|FEEDBACK TEXT|)@. \\
Item - \verb@(|CARD TO SORT|)@. \\
Options - 1. one red circle, 2. two green triangles, 3. three blue crosses, 4. four yellow stars \\
\end{tabbing}
\end{promptbox}

\begin{promptbox}[Data Format 5 (Field\textbackslash t\{\}. Answer\textbackslash t\{\}) - WCST]
\begin{tabbing}
Feedback	\= \verb@(|FEEDBACK TEXT|)@. \\
Item	\= \verb@(|CARD TO SORT|)@. \\
Options	\= 1. one red circle, 2. two green triangles, 3. three blue crosses, 4. four yellow stars \\
\end{tabbing}
\end{promptbox}

\begin{promptbox}[Data Format 6 (FIELD- \{\} \textbackslash n ANSWER- \{\}) - WCST]
\begin{tabbing}
FEEDBACK- \verb@(|FEEDBACK TEXT|)@ \\
ITEM- \verb@(|CARD TO SORT|)@ \\
OPTIONS- 1. one red circle, 2. two green triangles, 3. three blue crosses, 4. four yellow stars \\
\end{tabbing}
\end{promptbox}

\begin{promptbox}[Data Format 7 (field:: \{\} -- answer:: \{\}) - WCST]
\begin{tabbing}
feedback:: \verb@(|FEEDBACK TEXT|)@ \= -- \\
item:: \verb@(|CARD TO SORT|)@ \= -- \\
options:: 1. one red circle, 2. two green triangles, 3. three blue crosses, 4. four yellow stars \\
\end{tabbing}
\end{promptbox}

\begin{promptbox}[Data Format 8 (field - \{\}\texttt{ ,  }answer - \{\}) - WCST] 
\begin{tabbing}
feedback - \verb@(|FEEDBACK TEXT|)@ , \\
item - \verb@(|CARD TO SORT|)@ , \\
options - 1. one red circle, 2. two green triangles, 3. three blue crosses, 4. four yellow stars \\
\end{tabbing}
\end{promptbox}

\begin{promptbox}[Data Format 9 (Field \textbackslash n\textbackslash t\{\} \textbackslash n Answer \textbackslash n\textbackslash t\{\}) - WCST]
\begin{tabbing}
Feedback \\
	\= \verb@(|FEEDBACK TEXT|)@ \\
Item \\
	\= \verb@(|CARD TO SORT|)@ \\
Options \\
	\= 1. one red circle, 2. two green triangles, 3. three blue crosses, 4. four yellow stars \\
\end{tabbing}
\end{promptbox}

\begin{promptbox}[Data Format 10 (Field - \{\}\textbackslash n Answer - \{\}) - WCST]
\begin{tabbing}
Feedback - \verb@(|FEEDBACK TEXT|)@ \\
Item - \verb@(|CARD TO SORT|)@ \\
Options - 1. one red circle, 2. two green triangles, 3. three blue crosses, 4. four yellow stars \\
\end{tabbing}
\end{promptbox}

\end{document}